\pdfoutput=1
\documentclass[11pt]{article}
\usepackage{acl}
\usepackage{times}
\usepackage{latexsym}
\usepackage[T1]{fontenc}
\usepackage[utf8]{inputenc}
\usepackage{microtype}
\usepackage{inconsolata}
\usepackage{graphicx}
\usepackage{booktabs}
\usepackage{amsfonts}
\usepackage{amsmath}
\usepackage{mathtools}
\usepackage{amssymb}
\usepackage{bbm}
\usepackage{float}
\usepackage{array}
\usepackage{algorithm2e}
\usepackage{algpseudocode}
\usepackage{hyperref}
\usepackage{relsize}
\usepackage{multirow}
\usepackage{enumitem}
\usepackage{tabularx}
\usepackage{listings}
\usepackage{xcolor}
\usepackage{changepage}
\usepackage{color}
\usepackage{colortbl}

\usepackage{multicol}
\usepackage{minitoc} % Import the minitoc package
\usepackage[toc,page]{appendix} 

\usepackage{amssymb}% http://ctan.org/pkg/amssymb
\usepackage{pifont}% http://ctan.org/pkg/pifont
\newcommand{\xmark}{\ding{55}}%

\definecolor{darkgreen}{HTML}{548235}
\definecolor{darkred}{HTML}{C00000}
\definecolor{darkerblue}{HTML}{240394}
\definecolor{darkblue}{HTML}{2e75B6}
\definecolor{darkyellow}{HTML}{BF9000}
\definecolor{darkpurple}{HTML}{7030A0}

\newcommand{\methodname}{DiVE\xspace}

\newcommand{\code}[1]{\texttt{#1}}
\newcommand{\cmd}[1]{\textcolor{darkgreen}{\code{#1}}}
\newcommand{\minihack}[1]{\textcolor{darkred}{\code{#1}}}

\title{Enhancing Agent Learning through World Dynamics Modeling}

\author{
Zhiyuan Sun$^1$\thanks{\:\:Equal contribution.} \thanks{\:\:zhiyuan.sun@umontreal.ca}, Haochen Shi$^1$\footnotemark[1] \\
\textbf{Marc-Alexandre Côté$^2$, Glen Berseth$^{1, 3}$, Xingdi Yuan$^2$\thanks{\:\:Equal advising.}, Bang Liu$^{1, 3}$\footnotemark[3]} \\
$^1$ Université de Montréal \& Mila, Montréal, Canada \\
$^2$ Microsoft Research, Montréal, Canada \\ 
$^3$ Canada CIFAR AI Chair\\ 
}

\begin{document}
\maketitle 

% abstract section
\begin{abstract}
Large language models (LLMs) have been increasingly applied to tasks in language understanding and interactive decision-making, with their impressive performance largely attributed to the extensive domain knowledge embedded within them. However, the depth and breadth of this knowledge can vary across domains. Many existing approaches assume that LLMs possess a comprehensive understanding of their environment, often overlooking potential gaps in their grasp of actual world dynamics. To address this, we introduce \textbf{Di}scover, \textbf{V}erify, and \textbf{E}volve (\textbf{\methodname}), a framework that \textbf{discovers} world dynamics from a small number of demonstrations, \textbf{verifies} the accuracy of these dynamics, and \textbf{evolves} new, advanced dynamics tailored to the current situation. Through extensive evaluations, we assess the impact of each component on performance and compare the dynamics generated by \methodname to human-annotated dynamics. Our results show that LLMs guided by \methodname\ make more informed decisions, achieving rewards comparable to human players in the Crafter environment and surpassing methods that require prior task-specific training in the MiniHack environment\footnote{\:\:The code is available at \url{https://github.com/ZhiyuuanS/DiVE}}.

% ZY.09.26.2024: add the MiniHack's performance on the abstract.
\end{abstract}

\section{Introduction}
\begin{figure}[t!]
\centering
\includegraphics[width=0.5\textwidth]{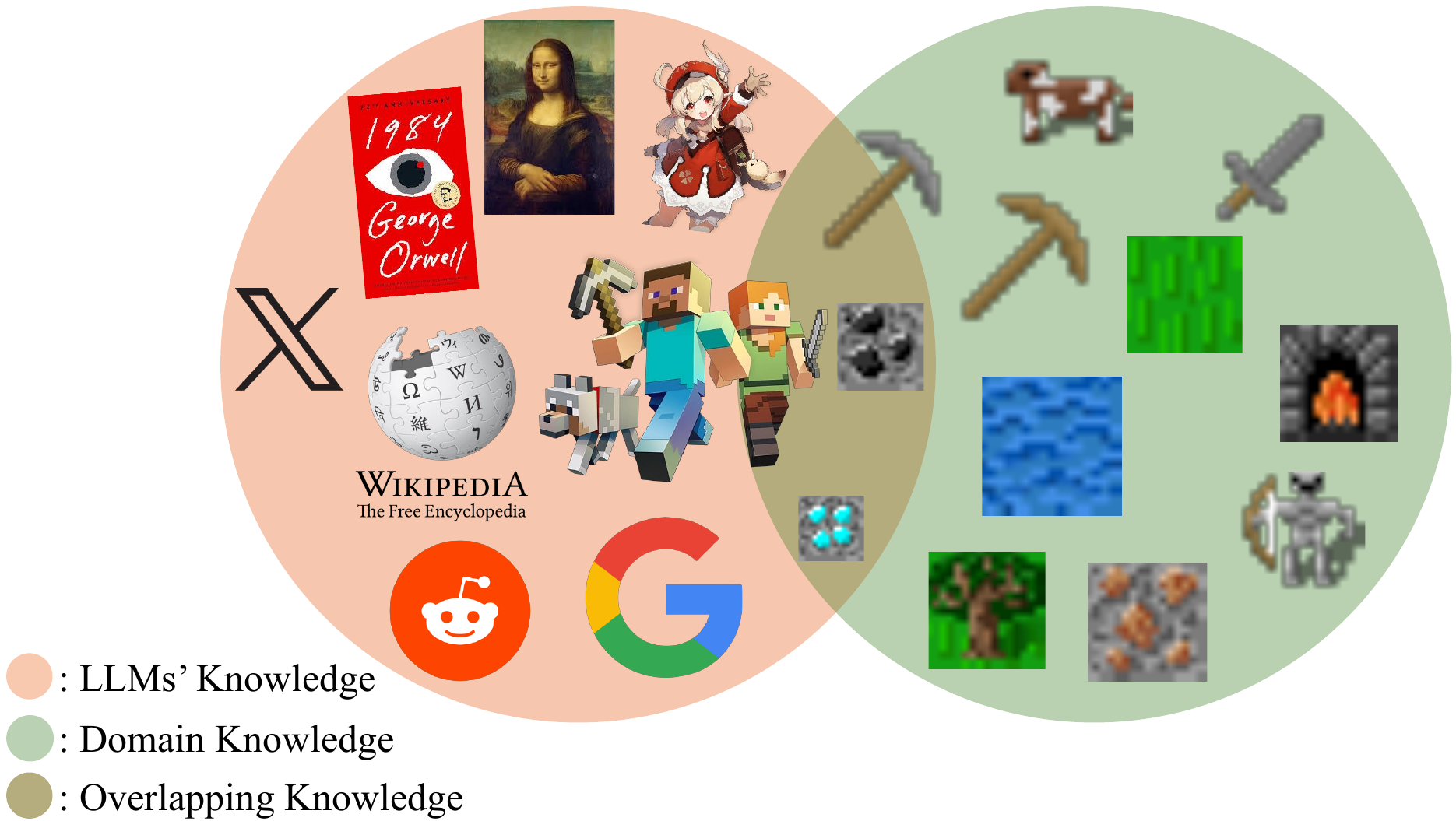}
\caption{The knowledge gap between LLMs and downstream domains. Although LLMs have a broad understanding of the world, they may struggle to grasp the complex dynamics of specific downstream domains.}
\label{fig:motivation}
\vspace{-1em}
\end{figure}

By absorbing internet-scale knowledge autoregressively, large language models (LLMs) develop a broad understanding of the world~\cite{achiam2023gpt, team2023gemini, brown2020language, touvron2023llama}. This understanding enables them to perform well across a variety of tasks~\cite{brown2020language, huang2022language, yao2022react, yao2024tree}. However, achieving this level of comprehension requires a training dataset that is diverse, precise, and in-depth to cover essential domain information. Otherwise, a knowledge gap may emerge between LLMs and the target domain, as illustrated in Figure \ref{fig:motivation}.

Since LLMs tend to follow the most common patterns in the dataset~\cite{gunasekar2023textbooks} and are prone to hallucination, it is crucial that the necessary information not only appears frequently on the internet but is also reliable~\cite{dziri-etal-2022-origin}. For tasks and environments that are newly emerging~\cite{hafner2021benchmarking, samvelyan2021minihack}, these requirements are often difficult to meet due to the noisy nature of internet data, leading to potential knowledge gaps.

While LLMs may have a general understanding of a domain, optimal decision-making requires in-depth, state-specific knowledge. Providing LLMs with such tailored information enhances their grasp of both the environment and the current state. For example, mastering Go involves not just knowing the rules but also applying strategies for specific board states. This specialized knowledge varies across domains and states, making offline collection impractical.

To address these challenges, we propose \textbf{Di}scover, \textbf{V}erify, and \textbf{E}volve (\textbf{\methodname}). Building on the concept of the World Model \cite{ha2018worldmodels}, \methodname not only discovers and verifies primitive world dynamics from demonstrations but also evolves state-specific knowledge for the downstream domain. By providing LLMs with this comprehensive set of dynamics, \methodname bridges the knowledge gap between LLMs and the downstream domain, thereby enhancing their decision-making abilities.

\methodname consists of three distinct components:
The \textbf{Discoverer}: this component iteratively uncovers the environment's dynamics from demonstrations using a curriculum learning approach.
The \textbf{Verifier}: this component eliminates unreliable dynamics caused by LLMs' tendency to hallucinate.
The \textbf{Evolver}: this component reasons through in-depth, state-specific strategies tailored to the current situation based on the learned dynamics.

In the Crafter environment \cite{hafner2021benchmarking} and the MiniHack environment~\cite{samvelyan2021minihack}, \methodname learns comprehensive and reliable dynamics from demonstrations, guiding the agent's decision-making process by evolving in-depth strategies. This enables the agent to outperform all baselines, achieving rewards comparable to human players in the Crafter environment and surpassing the performance of methods requiring task-specific training in the MiniHack environment. To gain deeper insights into \methodname's behavior, we provide both quantitative and qualitative analyses.

In summary, our primary contribution is a framework that learns world dynamics from demonstrations, guiding LLMs in the decision-making process by online evolving contextual strategies. This approach bridges potential knowledge gaps, resulting in a more optimal decision-making process for LLMs.

% ZY.09.26.2024: Reformat the introduction section to reduce the length.

\section{The Knowledge Gap}
% ZY.09.27.2024. Change the section name from problem formulation to The Knowlegde Gap
\begin{figure*}[t!]
\centering
    \includegraphics[width=0.48\linewidth]{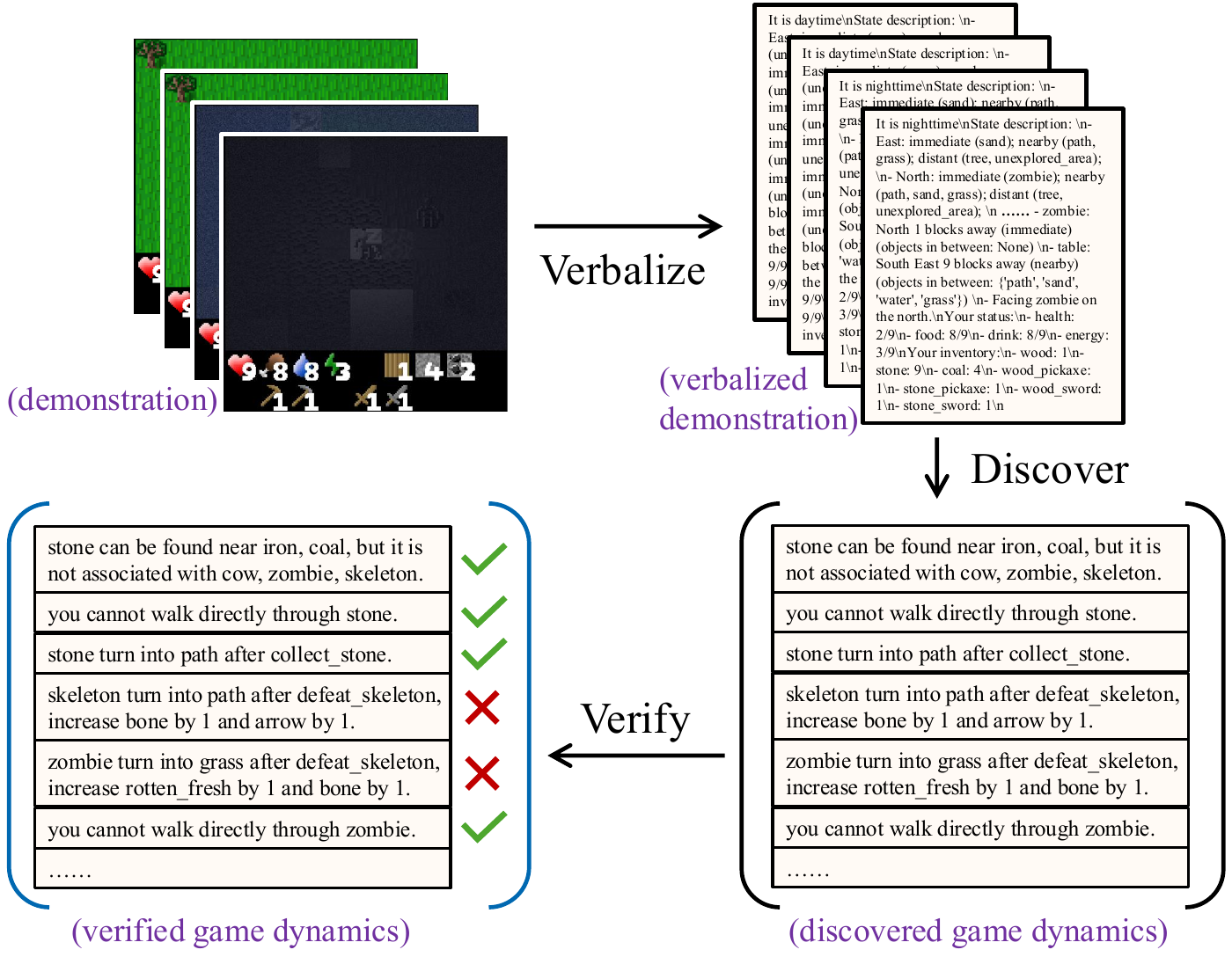}
    \hspace{1em}
    \includegraphics[width=0.48\linewidth]{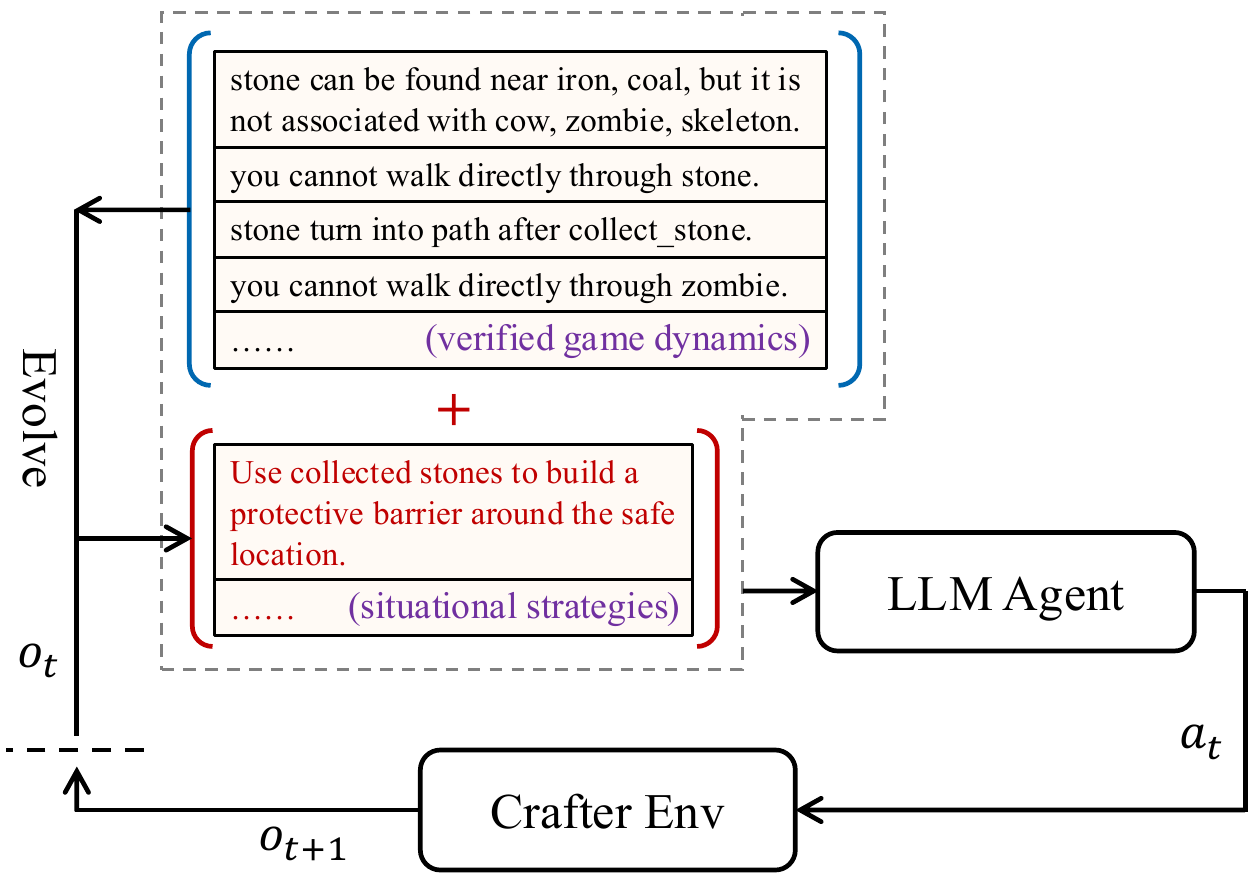}
    \caption{Overall pipeline of \methodname. \textbf{Left}: Learning basic game dynamics from offline demonstrations (Section~\ref{sec:method:offline}). We want to highlight the incorrect game dynamics being identified by the Verifier (labeled by \textcolor{darkred}{$\times$}), they are evidence of the LLMs hallucinate false facts perhaps because of memorizing Minecraft data. \textbf{Right}: Learning situational strategies from online interactions (Section~\ref{sec:method:online}). 
    For simplicity, we omit the verbalization process in the right figure.} 
    \label{fig:general}
    \vspace{-0.5em}
\end{figure*}

In this work, we consider a setting where a pre-trained LLM-based agent is employed to solve tasks in downstream domains. Conceptually, we define $\mathcal{K}_{\text{LLM}}$ as the set of knowledge embedded in the LLMs through their training process, and $\mathcal{K}_{\text{target}}$ as the universal set of knowledge relevant to the downstream domains.

To ensure the effectiveness of the LLMs, we aim for $\mathcal{K}_{\text{relevant}}$, the subset of $\mathcal{K}_{\text{LLM}}$ that is relevant to $\mathcal{K}_{\text{target}}$, to cover the broadest possible scope. Additionally, we seek for $\mathcal{K}_{\text{relevant}}$ to contain more reliable knowledge $\mathcal{K}_{+}$ than inaccurate knowledge $\mathcal{K}_{-}$, where $\mathcal{K}_{\text{relevant}} = \mathcal{K}_{+} \cup \mathcal{K}_{-}$.
Thus, we define three desirable properties to guide our system design:

\begin{itemize}[leftmargin=*]
    \item \textbf{Recall} $R = \frac{|\mathcal{K}_{+}|}{|\mathcal{K}_{\text{target}}|}$ measures the extent to which the knowledge required to solve tasks in the target domain is covered by the LLMs. A low recall typically indicates a significant knowledge gap between the LLMs' training data and the target domain.
    \item\textbf{Precision} $P = \frac{|\mathcal{K}_{+}|}{|\mathcal{K}_{\text{relevant}}|}$ measures the accuracy of the LLMs' knowledge when applied to the target domain. An example of inaccurate dynamics is shown in Figure~\ref{fig:general} (left), where the LLMs hallucinate that defeating a skeleton will drop items, which is not true in Crafter. This may be due to the LLMs memorizing Minecraft-specific data.
    \item\textbf{Depth} $D$ measures the abstraction levels of knowledge representations. Knowledge can be represented at varying levels of abstraction, from basic game rules to higher-level strategies.
\end{itemize}

We recognize that precisely quantifying the knowledge overlap between LLMs and a downstream domain is challenging. In Section~\ref{sec:method}, we provide a mathematical analysis showing how \methodname improves across all three dimensions. In Section~\ref{sec:exp}, we measure the knowledge overlap to demonstrate that \methodname effectively learns useful game dynamics and, to some extent, bridges the knowledge gap.

\section{\methodname: Discover, Verify, and Evolve}
\label{sec:method}
% ZY.09.27.2024. Change the section name from Method to DiVE
In an ideal scenario, one could bridge the knowledge gap by fine-tuning the LLM to adapt to the target domain. However, this approach is often less practical due to its reliance on large amounts of annotated data and significant computational overhead~\cite{hu2021lora, zheng2024fine, pmlr-v202-carta23a, ouyang2022training}.
Our framework, \methodname, is designed to address the knowledge gap while considering all three desirable properties—recall, precision, and depth—without requiring extensive data collection from the target domain. It is a prompt-based method that learns world dynamics $\mathcal{W}$ directly from the environment.
% ZY.09.27.2024. Remove the math symbol $\tau$ since it is not used in later sections, and improve the clarity.

As shown in Figure~\ref{fig:general},  \methodname is initially bootstrapped using a small set of human demonstration trajectories $H$, each consisting of observations $o_t$, actions $a_t$, and rewards $r_t$ at each timestep $t$. We then transform each observation $o_t$ into the language space as $\tilde{o}_t$ using a Verbalizer, resulting in trajectories represented by $\tilde{o}_t$, $a_t$, and $r_t$ at each timestep $t$. 
Subsequently, the \textbf{Discoverer} extracts a set of world dynamic candidates, $\tilde{\mathcal{W}} = \{\tilde{W}_{+}, \tilde{W}_{-}\}$, from human demonstrations $H$, where $\tilde{W}_{+}$ and $\tilde{W}_{-}$ represent the correct and inaccurate world dynamic sets, respectively.

Empirically, we find that the inclusion of $\tilde{W}_{-}$ in $\tilde{\mathcal{W}}$ is often unavoidable, either due to LLM's difficulties in extracting meaningful knowledge from trajectory data or its tendency to hallucinate. 
To address this, we employ the \textbf{Verifier} to filter out potentially invalid and conflicting world dynamic candidates from $\tilde{\mathcal{W}}$, leaving only the valid dynamics $\mathcal{W}$.
Lastly, we use the \textbf{Evolver}, designed to derive advanced dynamics $\mathcal{I}$ tailored to the verbalized observation $\tilde{o}_t$ based on the filtered world dynamics $\mathcal{W}$.

The final decision-making process on primitive actions $a_t \in \mathcal{A}$ is hierarchically decomposed as planning tasks on sub-goals $SG$, sub-tasks $ST$, and actions $\mathcal{A}$. The planning procedure is further guided by both $\mathcal{W}$ and $\mathcal{I}$. In cases where $\mathcal{W} \neq \varnothing$, Recall, Precision and Depth are guaranteed to increase as formulated below: 
% ZY.09.27.2024. $R$, $P$, and $D$ are not defined, as we only use them here. Perhaps we can use their full names instead for clarity.
\begin{equation}
\begin{aligned}
    &\textbf{Recall}: \frac{|\mathcal{K}_{+}|}{|\mathcal{K}_{\text{target}}|} \xRightarrow[H]{\text{Discoverer}} \frac{|\mathcal{K}_{+}| + |\mathcal{W}|}{|\mathcal{K}_{\text{target}}|} \\
    &\textbf{Precision}: \frac{|\mathcal{K}_{+}|}{|\mathcal{K}_{\text{relevant}}|} \xRightarrow[\tilde{\mathcal{W}}, H]{\text{Verifier}} \frac{|\mathcal{K}_{+}| + |\mathcal{W}|}{|\mathcal{K}_{\text{relevant}}| + |\mathcal{W}|} \\
    &\textbf{Depth}: \varnothing \xRightarrow[\text{Verifier}]{\text{Discoverer}} \mathcal{W} \xRightarrow[\mathcal{W }, H]{\text{Evolver}} \mathcal{I} \cup \mathcal{W}
\end{aligned}
\notag
\end{equation}

The \methodname\ framework can be divided into two stages: an offline dynamics learning stage and an online strategy learning stage.

\subsection{Offline Dynamics Learning} 
\label{sec:method:offline}
The offline dynamics learning procedure aims to bridge the knowledge gap between LLMs' understanding and the basic dynamics of downstream domains by learning the world dynamics $\mathcal{W}$ as a prior for the decision-making process.
Instead of relying on human-authored game manuals or handbooks to extract world dynamics, as in \cite{wu2024spring, wu2024read}, which are not only difficult to obtain in many real-world scenarios but also often lack critical details (as demonstrated in Table~\ref{tab:dynamics_comparision}), we propose learning world dynamics $\mathcal{W}$ directly from experiences $H$, which are more accessible and provide richer information.

\paragraph{Hierarchical Curriculum Learning} 
Given the varying complexities of learning the dynamics of different elements in downstream domains, we adopt a curriculum learning approach~\cite{bengio2009curriculum}. Our method follows a sequential learning strategy that progresses from simpler to more complex dynamics, thereby enabling more effective learning. Specifically, we propose a method for learning the dynamics of each element within the task decomposition hierarchy, denoted as $TD = \{\mathcal{A} \cup O, ST, SG\}$, where $O$ represents the set of objects in downstream domains.

Our approach begins with elements of lower abstraction, such as actions $a \in \mathcal{A}$ and objects $o \in O$, and gradually progresses to higher-level elements, such as sub-tasks $st_i \in ST$. The sub-tasks $ST$ are represented as nodes in the achievement graph $\mathcal{G} = (V, E)$ within the downstream domains, i.e., $ST = V$. Finally, we transition to the subgoals $sg_i \in SG$. The subgoal sequence $SG = [sg_1, sg_2, \ldots]$ is an ordered list used to unlock achievements in the achievement graph $\mathcal{G}$, where $SG = \text{TopologicalSort}(\mathcal{G})$, and each $sg_i$ corresponds to a vertex in $V$.

We leverage the Discoverer to extract this order from human demonstrations $H$. Achieving a sub-goal $sg_i$ may involve completing several sub-tasks multiple times. This approach ensures a logical progression through tasks, thereby enabling a deeper understanding and integration of downstream domain dynamics.

\paragraph{Discoverer} 
The Discoverer is designed to identify dynamic candidates $\tilde{\mathcal{W}}$ related to elements within the task decomposition hierarchy $TD$. A single dynamics discovery step for an element $E \in TD$ involves the following three main steps:
\begin{enumerate}[leftmargin=*]
    \item \textbf{Construction of the Semantic Experience Bank}: For each element $E$, we construct a semantic experience bank $B^{E}$ using demonstrations $H$. This bank stores experiences that are transformed from $H$ into a suitable granularity for analyzing dynamics related to $E$. The transformation process involves chunking and summarizing the verbalized demonstrations to capture essential semantic details.
    \item \textbf{Sampling of Relevant Experiences}: For each attribute of an instance $e \in E$, a subset of experiences $B_e^E$ that are relevant to the instance $e$ is sampled from $B^E$. 
    \item \textbf{Identification of Dynamic Candidates}: A dynamic candidate $\tilde{w}$ is identified from the subset $B_e^E$ by recognizing patterns that are consistent across all experiences within $B_e^E$. 
\end{enumerate}

The action-level semantic experience bank, denoted as $B^\mathcal{A}$, stores transition tuples derived from verbal demonstrations and is represented as: $B^\mathcal{A} = \{\{\tilde{o}_t, a_t, \tilde{o}_{t+1}\}_i\}_{i=1}^{|B^\mathcal{A}|}$. 
Similarly, the object-level semantic experience bank, denoted as $B^O$, gathers individual observations contain an specific object and is represented as: $B^O = \{\tilde{o}_i\}_{i=1}^{|B^O|}$. 
The sub-task-level semantic experience bank, denoted as $B^{ST}$, aggregates trajectory segments representing the completion of sub-tasks and is formatted as: $B^{ST} = \{\{\tilde{o}_t, \ldots, a_{t_{st}}, \tilde{o}_{t_{st}+1} \}_i\}_{i=1}^{|B^{ST}|}$, where $t_{st}$ denotes the timestep at which a sub-task $st \in ST$ is completed.
For sub-goals, the sub-goal-level experience bank, denoted as $B^{SG}$, records sequences of sub-tasks that culminate in the completion of sub-goals, expressed as: $B^{SG} = \{\{st_t, \ldots, st_{t_{sg}}\}_i\}_{i=1}^{|B^{SG}|}$, where $t_{sg}$ is the timestep at which the sub-goal $sg$ is achieved.

For action-level dynamics learning, the relevant experiences, denoted as $B_a^{\mathcal{A}}$, are compiled by randomly sampling transition tuples from $B^{\mathcal{A}}$ where the action $a$ has been successfully executed.
A similar approach is employed for dynamics learning of other elements within the task decomposition hierarchy $TD$.

For action-level dynamics discovery, we identify the prerequisites and outcomes of each action (e.g., \cmd{The action MakeWoodPickaxe requires 1 wood}). For object-level dynamics, we focus on co-occurrence relationships between objects and their temporal patterns. The attribute set for a sub-task generally encompasses the necessary steps for completion, as well as its prerequisites, outcomes, and termination conditions. In contrast, the primary attribute of interest for a sub-goal is its correct position within the sub-goal sequence $SG$.

\paragraph{Verifier}
Dynamic discovery processes are vulnerable to noise from various sources, including confounders, hallucinations by the LLMs, and difficulties in the LLMs' ability to extract meaningful insights from trajectory data.
To address these challenges, we introduce a dynamic verifier designed to filter out noisy dynamic candidates from $\tilde{\mathcal{W}}$. 
For each dynamic candidate $w \in \tilde{\mathcal{W}}$, the verifier begins by sampling a subset of relevant semantic experiences, denoted as $B_e^E$, from the corresponding semantic experience bank $B_E$. Here, $w$ represents a dynamic candidate associated with a specific attribute of the instance $e \in E$, where $E \in \{\mathcal{A} \cup O, ST, SG\}$ corresponds to an element of the task decomposition hierarchy $TD$.
The verification of $w$ is conducted as follows: $w$ is deemed inaccurate and filtered out if it does not consistently hold across experiences within $B_e^E$ or if it conflicts with any established dynamics. The dynamics that pass this verification process are classified as verified dynamics and are denoted as $\mathcal{W}$.

\subsection{Online Strategy Learning}
\label{sec:method:online}
To effectively incorporate the learned world dynamics $\mathcal{W}$ into the downstream domains, we deploy an LLM-based agent defined by $\pi: \mathcal{S} \times \mathcal{W} \rightarrow \mathcal{P}(\mathcal{A})$. Here, $\mathcal{S}$ represents the state space, $\mathcal{A}$ denotes the action space, and $\mathcal{P}$ symbolizes the probability distribution over the action space.   
Instead of directly mapping the world dynamics $\mathcal{W}$ and the current state observation $o_t$ to the action $a_t$, we tackle the challenge of long-horizon planning by integrating an online strategy learning method. This approach decomposes the planning process into three distinct stages: sub-goal planning, sub-task planning, and action planning.

\paragraph{Sub-goal Planning}
Given that the sub-goal sequence $SG = [sg_1, sg_2, \ldots]$ is derived from human demonstrations $H$ and treated as a fixed sequence, we utilize a straightforward heuristic for sub-goal planning. When a sub-goal is completed, the current sub-goal is updated to the first uncompleted sub-goal in $SG$.
\paragraph{Sub-task Planning}
For a given current sub-goal $sg_i$, we have developed an LLM-based sub-task planner. This planner evaluates and ranks all sub-tasks $st \in ST$ based on the learned world dynamics $\mathcal{W}$, the verbalized current observation $\tilde{o}_t$, and the most recently executed sub-task $st_{t-1}$. The highest-ranked sub-task is then designated as the current sub-task $st_t$. To ensure accurate execution, the completion of a sub-task $st$ is contingent upon satisfying its specific termination condition. This condition is verified by querying an LLM using the current verbalized observation, the observation at the time the sub-task began, and the termination conditions of the current sub-task.

\paragraph{Learning Strategies}
In addition to learning the fundamental rules of the downstream domains, we also focus on developing advanced game-playing strategies based on these dynamics. Unlike world dynamics learning, the strategy space is often too expansive for exhaustive exploration. To address this challenge, we propose evolving the dynamics into strategies, denoted as $\mathcal{I}$, using an online learning approach.

This method reduces the search space by conditioning not only on the dynamics $\mathcal{W}$, but also on the verbalized current observation $\tilde{o}_t$ and the sub-task $st_i$. This targeted approach enables the generation of strategies that are more contextually grounded and responsive to current game scenarios compared to those developed through offline methods.
To support this process, we have designed an LLM-based Evolver that generates strategy candidates $\tilde{\mathcal{I}}$ by applying deductive reasoning to the learned dynamics $\mathcal{W}$.
Specifically, the Evolver derives strategy candidates using rules of inference, such as modus ponens. These strategy candidates, denoted as $\tilde{\mathcal{I}}$, are then evaluated for validity and ranked based on their utility by an LLM-based critic. Finally, the valid and useful candidates are incorporated into the situational strategy set $\mathcal{I}$.

\paragraph{Action Planning}
The final action selection process is executed in two main steps:
\begin{enumerate}[leftmargin=*]
\item Invalid Action Masking: This step involves filtering out actions that are infeasible under the current situation, based on the verified dynamics $\mathcal{W}$ and current verbalized observation $\tilde{o}_t$.
\item Action Selection: From the set of valid actions, a specific primitive action $a$ is chosen based on multiple factors: the current sub-task $st_i$, the verbalized current observation $\tilde{o}_t$, the world dynamics $\mathcal{W}$, a windowed history of previously planned actions and observations, and the derived strategies $\mathcal{I}$.
\end{enumerate}

% ZY.09.27.2024. Split the Experiment section into Experiment Setup and Experiment Results
\section{Experiment Setup}
To demonstrate the effectiveness of \methodname\ in bridging the knowledge gap, we evaluate its performance within the Crafter and MiniHack environments. For a fair comparison, all LLM-based agents, including ours, utilize the GPT-4o model and the same environment seed in the Crafter setting. Further details on the setups can be found in Appendix~\ref{Appendix:minihack_setup} and~\ref{Appendix:crafter_setup}.

\subsection{Crafter}
Crafter \cite{hafner2021benchmarking} is an open-world survival game set on $64 \times 64$ grid-based maps, featuring a diverse array of materials such as \cmd{tree}, \cmd{stone}, and \cmd{coal}, as well as entities including \cmd{cow}, \cmd{zombie}, and \cmd{skeleton} semi-randomly spawn on the maps. The games include an achievement graph with 22 unique achievements across 7 levels. The agent perceives its surroundings through a local $7 \times 9$ observation window and maintains awareness of its status within the game environment.

The text description generated by the verbalizer includes: the nearest object of each type within the accumulated observations, the objects situated between these nearest objects, the objects in each direction, as well as the agent's current inventory and status. An example of the verbalization process is provided in Appendix~\ref{sec:verbalizer_example}.

The agent is evaluated using two primary metrics: reward and score. Agents receive a \( +1 \) reward for each new achievement unlocked (e.g., \cmd{make wood pickaxe}, \cmd{place furnace}) and a \( \pm 0.1 \) reward for every health point gained or lost. The score is calculated by aggregating the success rates $s_i$ for each achievement and is formulated as:
\begin{equation}
    \footnotesize
    S \doteq \exp \left( \frac{1}{N} \sum_{i=1}^N \ln \left(1+s_i\right) \right) - 1.
    \notag
\end{equation}

\vspace{2em}
\noindent We compare \methodname against the following baselines:
\vspace{-0.5em}
\begin{itemize}
    \item \textbf{LLM-based approaches:} SPRING~\cite{wu2024spring}, ELLM~\cite{du2023guiding}, and Chain-of-Thought (CoT)\cite{wei2022chain}. 
    % \vspace{-0.5em}
    \item \textbf{Reinforcement Learning (RL) approaches:} DreamerV3\cite{hafner2023mastering}, PPO~\cite{schulman2017proximal}, and AD~\cite{moon2024discovering}.
    % \vspace{-2em}
    \item \textbf{Human Players:} Expert performance on the Crafter environment.
\end{itemize}

\subsection{MiniHack}
MiniHack~\cite{samvelyan2021minihack} is a grid-based environment built on the video game NetHack~\cite{kuettler2020nethack}. Unlike the Crafter environment, it supports the creation of tasks that target specific agent capabilities. In our work, we focus on the Skill Acquisition Tasks subset, which evaluates the agent's ability to leverage the rich diversity of NetHack's objects, monsters, and dungeon features, as well as their interactions. These tasks introduce complex world dynamics, where actions are factorized autoregressively and require executing a sequence of follow-up actions for the initial action to produce the desired effect.

We use the Lava Crossing, Wand of Death, and Quest tasks as testbeds to evaluate \methodname. 
We provide detailed descriptions of these task in Appendix~\ref{appendix:des_files}. The agent is rewarded for unlocking achievements within each task. To facilitate interaction with LLMs, we use the NLE language wrapper~\cite{goodger2023nethack} to verbalize both observations and actions from the environment.

We compare \methodname\ against SSO~\cite{nottingham2024sso} and Reflexion~\cite{shinn2024reflexion}, both of which require prior training on the tasks. 
Following the official guidelines, we run each method for 30 iterations, evaluating performance after every 10 iterations by attempting the task 10 times using a fixed set of skills or reflections.
The final performance is determined based on the results from the last 10 evaluation attempts.

\section{Experimental Results}
\begin{table}[t!]
\begin{center}
\small
\begin{tabular}{p{2.2cm} @{\hskip 1.34cm} c  c}
\toprule
\multicolumn{1}{l}{Method} & \multicolumn{1}{c}{Score} & \multicolumn{1}{c}{Reward} \\
\toprule
Human Experts & $50.5\pm6.8\%$ & $14.3\pm2.3$ \\
\midrule
DiVE & $\mathbf{35.9\pm3.2\%}$ & $\mathbf{14.5\pm2.4}$ \\
SPRING$^{*}$ & $8.2\pm2.3\%$ & $6.9\pm1.8$ \\
CoT & $1.3\pm0.3\%$ & $2.5\pm0.5$\\
AD & $21.79 \pm 1.4\%$ & $12.6\pm0.3$ \\
ELLM &  N/A & $6.0 \pm 0.4$ \\
DreamerV3 & $14.5 \pm 1.6 \%$ & $11.7 \pm 1.9$\\
PPO & $4.6 \pm 0.3 \%$ & $4.2 \pm 1.2$  \\
Random & $1.6 \pm 0.0\%$ &  $2.1 \pm 1.3$ \\
\bottomrule
\end{tabular}
\end{center}
\vspace{-0.5em}
\caption{Performance comparison of \methodname\ against baseline models in the Crafter environment. Methods with $^{*}$ indicate that they were obtained using the official code implementation, executed with the same five random seeds and model configurations as used for \methodname.}
\label{tab:overall}
\vspace{-1em}
\end{table}

% \eric{Did we say how many seeds we are using? If not, either say it here and table 2 caption, or in section 5.1.} \zhiyuan{added seed number}

We evaluate the performance of \methodname\ in the Crafter and MiniHack environments. In Section~\ref{sec:exp:result}, we present the overall results to demonstrate \methodname's effectiveness in bridging the knowledge gap in these tasks. Section~\ref{sec:exp:quantitative} provides a detailed analysis of the contributions of individual components through controlled experiments. Then, in Section~\ref{sec:exp:analysis}, we evaluate the effectiveness of the dynamics learned by \methodname\ in the Crafter environment. Finally, in Section~\ref{sec:exp:qualitative}, we further analyze the learned dynamics in both the Crafter environment and a modified MiniHack setting with altered dynamics.

\subsection{\methodname's Performance}
\label{sec:exp:result}
Table~\ref{tab:overall} and ~\ref{tab:minihack_performance} demonstrate that \methodname surpasses all other baselines in the Crafter and MiniHack environments. In the Crafter environment, \methodname exceeds the previous state-of-the-art (SOTA) LLM-based method, SPRING, by a substantial margin, achieving a 337.8\% relative improvement in score and a 110.1\% enhancement in reward. Additionally, \methodname also surpasses the prior SOTA RL-based approach, DreamerV3, with a 21.4\% absolute improvement in score and a 2.8 absolute increase in reward. Notably, \methodname achieves rewards comparable to human players using 10 demonstrations. 

In the MiniHack environment, with only a single demonstration, \methodname\ matches the performance of SSO and Reflexion (both of which require 30 iterations of training) on the Lava Crossing task, and outperforms both baselines on the Wand of Death and Quest tasks. Specifically, \methodname\ achieves a 68\% improvement over Reflexion on the Wand of Death task and a 30\% and 36\% improvement over SSO and Reflexion, respectively, on the Quest task.
\begin{table}[h!]
\begin{center}
\small
\begin{tabular}{p{0cm} @{\hskip 1cm} c  c c}
\toprule
\multicolumn{1}{l}{Method} & \multicolumn{1}{c}{Lava Crossing} & \multicolumn{1}{c}{Wand of Death} & \multicolumn{1}{c}{Quest}\\
\toprule
Reflexion     & 0.8$\pm$0.0    & 0.32$\pm$0.15 & 0.87$\pm$0.43     \\
SSO$^{*}$     & 0.8$\pm$0.0     & 0.52$\pm$0.31  & 0.91$\pm$0.29    \\ 
DiVE     & \textbf{0.8$\pm$0.0}   & \textbf{0.54$\pm$0.34}   & \textbf{1.18$\pm$0.38}  \\ 
\bottomrule
\end{tabular}
\end{center}
\vspace{-0.5em}
\caption{Comparison of \methodname\ performance against baseline models in the MiniHack environment.
 Results marked with $^{*}$ were obtained using the official code implementation and evaluated over 10 runs.}
\label{tab:minihack_performance}
\vspace{-1em}
\end{table}

\subsection{Contribution of Individual Components}
\label{sec:exp:quantitative}
\begin{table}[t!]
\centering
\small
\begin{tabular}{p{2.2cm} @{\hskip 1.35cm} c  c}
\toprule
Methods & Score & Reward \\
\hline
 \multicolumn{3}{l}{\textbf{Component analysis}} \\
 DiVE & $\mathbf{35.9\pm3.2\%}$ & $\mathbf{14.5\pm2.4}$ \\
 w/o E & $21.1\pm9.7\%$ & $11.3\pm4.3$ \\
 w/o V & $9.8\pm1.0\%$ & $10.1\pm0.7$ \\
 w/o V\&E & $11.5\pm4.9\%$ & $8.3\pm3.8$ \\
 w/o D\&V\&E & $0.9\pm0.1\%$ & $2.5\pm1.3$ \\
 CoT & $1.3\pm0.3\%$ & $2.5\pm0.5$ \\
 CoT + D\&V & $3.6\pm0.9\%$ & $3.9\pm2.3$  \\
\hline
 \multicolumn{3}{l}{\textbf{Dynamics from distinct sources}}\\
 DiVE & $\mathbf{35.9\pm3.2\%}$ & $\mathbf{14.5\pm2.4}$ \\
 w/o D\&V+S$^{\dag}$ & $15.7\pm5.3\%$ & $8.9\pm5.1$ \\
 w/o D\&V\&E+S$^{\dag}$ & $12.1\pm4.6\%$ & $8.7\pm3.0$ \\
 w/o D\&V\&E+H$^{\dag}$ & $34.2\pm2.8\%$ & $14.5\pm0.9$ \\
\bottomrule
\end{tabular}
\caption{Impact of different components on performance: Crafter. D, V, and E represent Discover, Verifier, and Evolver, respectively; S$^\dag$ refers to dynamics derived from the game manual, H$^\dag$ refers to human-annotated dynamics.}
\label{tab:ablation}
\vspace{-1em}
\end{table}
We conduct a series of ablation studies to clarify the contribution of each individual element to \methodname's overall performance. 
We report the results on Crafter in the first section of Table~\ref{tab:ablation}, and the results on MiniHack in Table~\ref{tab:minihack_different_component_performance}.
% As illustrated in the first section of Table~\ref{tab:ablation}, various variants of \methodname have been designed to assess the effectiveness of each component within the methodology. 

\paragraph{Crafter}
The significant performance gap between \methodname and its variant without the Evolver component empirically demonstrates the Evolver's effectiveness in developing gameplay strategies based on world dynamics $\mathcal{W}$, thereby enhancing the agent's overall proficiency in this environment. 
Similarly, the performance decline observed in the variant without the Verifier underscores the importance of formulating strategies $\mathcal{I}$ based on accurate world dynamics $\mathcal{W}$. Moreover, the further performance drop in the version lacking both the Verifier and Evolver components highlights their complementary roles—the Verifier ensures precision in capturing dynamics, while the Evolver focuses on strategy development.

The performance of \methodname without the Discoverer, Verifier, and Evolver components reverts to the CoT baseline, indicating that simply decomposing the task according to the hierarchy $\mathcal{H}$ without integrating domain knowledge $\mathcal{K}_{\text{target}}$ provides no performance benefit.
The substantial gap between CoT + D\&V and \methodname w/o E further demonstrates that an LLM-based agent struggles with long-horizon planning tasks in the absence of task decomposition, underscoring the importance of the decomposition hierarchy $\mathcal{H}$.

\paragraph{MiniHack}
Since MiniHack is built on the popular video game NetHack~\cite{goodger2023nethack}, LLMs already possess a certain level of understanding of the environment. For example, \methodname without the Discover, Verifier, and Evolver components (CoT baseline) can successfully solve the Lava Crossing task. However, as the complexity of the tasks increases, using LLMs alone is insufficient for solving more challenging scenarios like the Wand of Death and Quest tasks, highlighting a potential knowledge gap for these more intricate problems.

The primary performance gains of \methodname in the MiniHack environment can be attributed to the Discover component, which accurately identifies the underlying environment dynamics required for each task. Due to the embedded knowledge of LLMs about the MiniHack environment, they can identify these dynamics effectively, as shown by the minimal performance drop when \methodname\ is used without the Verifier. However, in the modified MiniHack setting described in Section~\ref{sec:exp:qualitative}, we highlight the critical role of the Verifier in maintaining reliability.
Additionally, since the tasks have a relatively short horizon, the dynamics uncovered by the Discover component are sufficient to complete them successfully. 

\begin{table}[t!]
\centering
\fontsize{8.5}{10}\selectfont
\begin{tabular}{p{1.6cm} @{\hskip -0.3cm} c c c}
\toprule
\multicolumn{1}{l}{Components} & \multicolumn{1}{c}{Lava Crossing} & \multicolumn{1}{c}{Wand of Death} & \multicolumn{1}{c}{Quest}\\
\toprule
\methodname&0.80$\pm$0.00&0.54$\pm$0.34&1.18$\pm$0.38\\
w/o E&0.72$\pm$0.24&0.52$\pm$0.30&1.16$\pm$0.48\\ 
w/o V&0.69$\pm$0.25&0.52$\pm$0.38&1.18$\pm$0.38\\ 
w/o V\&E&0.72$\pm$0.24&0.40$\pm$0.40&1.11$\pm$0.48\\
w/o D\&V\&E&0.72$\pm$0.17&0.20$\pm$0.00&0.41$\pm$0.29\\
\bottomrule
\end{tabular}
\vspace{-0.5em}
\caption{Impact of different components on performance: MiniHack. D, V, and E represent Discover, Verifier, and Evolver, respectively.}
\label{tab:minihack_different_component_performance}
\vspace{-1em}
\end{table}

\subsection{Evaluation of Learned Dynamics}
\label{sec:exp:analysis}
We investigate the performance of \methodname leveraging world dynamics derived from different sources in Crafter. 
As shown in the second section of Table~\ref{tab:ablation}, \methodname significantly outperforms variants that utilize the dynamics $S^{\dag}$ from the game manual \cite{wu2024spring}.
This performance improvement indicates that the learned dynamics $\mathcal{W}$ are more advantageous than $S^{\dag}$, likely because $S^{\dag}$ lacks certain beneficial details that are captured in $\mathcal{W}$.
The performance gap between methods using $S^{\dag}$ with and without the Evolver further highlights the importance of strategy evolution, whose effectiveness is closely tied to the quality of the underlying world dynamics.

In addition to dynamics learned from human demonstrations and game manual, we have explored a third source: human-annotated dynamics. The results show that \methodname performs comparably to the variants using human-annotated dynamics, demonstrating the robustness and effectiveness of \methodname's approach to dynamic learning.

As previously mentioned, it is difficult to quantify the desired properties because we cannot precisely measure the domain-relevant information $\mathcal{K}_{\text{relevant}}$ in LLMs or the exact amount of domain knowledge $\mathcal{K}_{\text{target}}$ required. However, by using human-annotated dynamics $H^{\dag}$ as a reference benchmark for $\mathcal{K}_{\text{target}}$, we can estimate the precision and recall of the learned dynamics $\mathcal{W}$, enabling us to effectively assess the progress of LLM-based dynamic learning. 
Specifically, we define recall as $R = \frac{|\mathcal{W} \cap H^{\dag}|}{|H^{\dag}|}$ and precision as $P = \frac{|\mathcal{W} \cap H^{\dag}|}{|\mathcal{W}|}$. 

\begin{figure}[t!]
\centering
\includegraphics[width=0.5\textwidth]{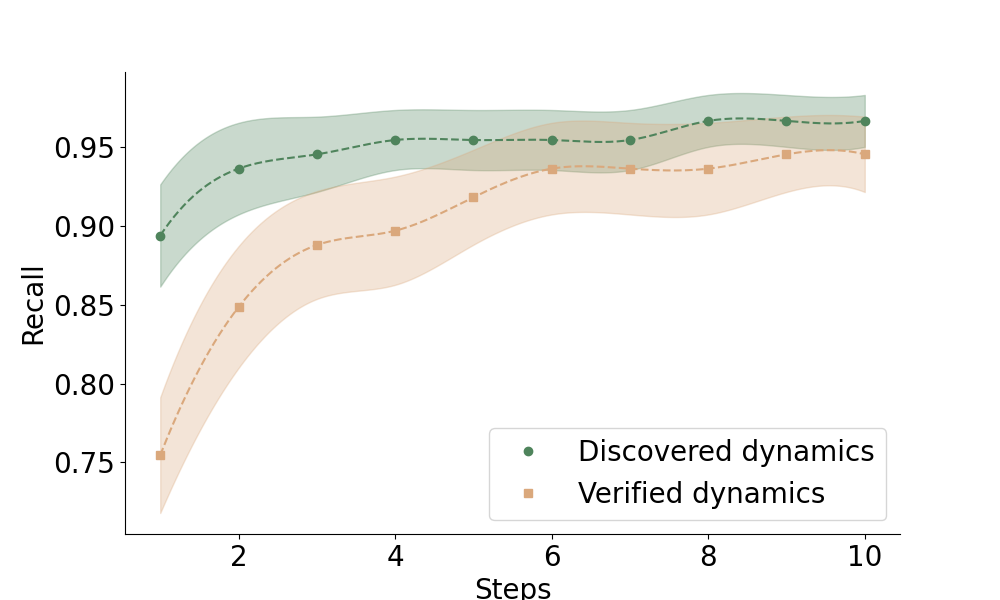}
\caption{Recall of learned dynamics over discovery steps, presented with mean and standard deviation, in the Crafter environment.}
\label{fig:dynamic_recall}
\vspace{-1em}
\end{figure}

As illustrated in Figure~\ref{fig:dynamic_recall}, both the discovered dynamics $\tilde{\mathcal{W}}$ and the verified dynamics $\mathcal{W}$ exhibit an increase in recall as the number of discovery steps progresses, indicating that the richness of the learned dynamics improves over time. Moreover, the narrowing gap in recall between $\tilde{\mathcal{W}}$ and $\mathcal{W}$ suggests that the Verifier effectively filters out 'noisy' dynamic candidates while preserving those that generalize across different trajectory segments.

\begin{figure}[t!]
\centering
\includegraphics[width=0.5\textwidth]{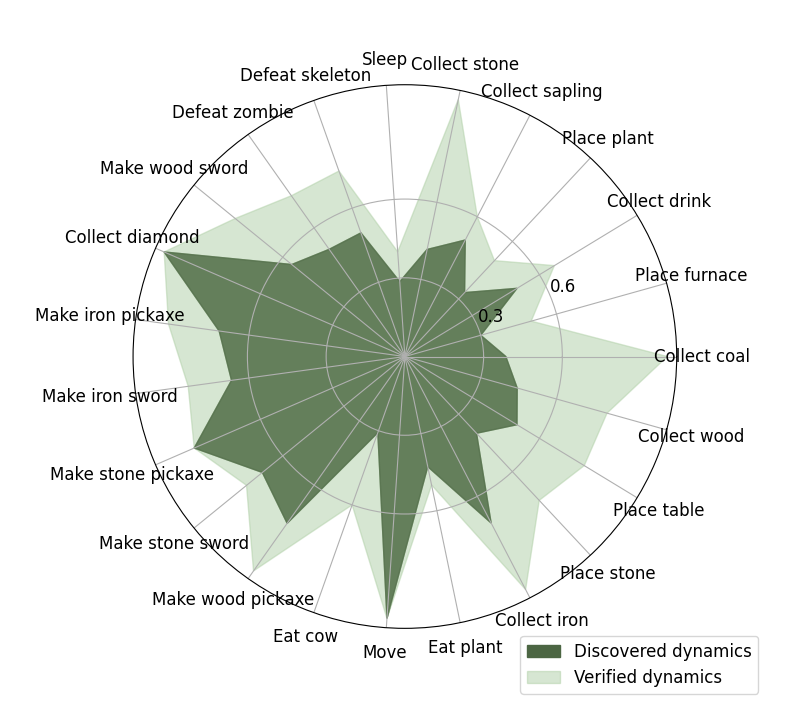}
\caption{Precision of learned dynamics before and after verification in the Crafter environment.}
\label{fig:dynamic_precision}
\vspace{-1em}
\end{figure}

To evaluate whether the Verifier preserves correct world dynamic candidates $\tilde{\mathcal{W}}{+}$ while filtering out unreliable ones $\tilde{\mathcal{W}}{-}$, we analyze the precision of both the discovered dynamics $\tilde{\mathcal{W}}$ and the verified dynamics $\mathcal{W}$. As shown in Figure~\ref{fig:dynamic_precision}, the precision of the verified dynamics consistently and significantly exceeds that of the discovered dynamics, demonstrating the Verifier's effectiveness in identifying and eliminating inaccurate candidates. This confirms the Verifier's role in enhancing the reliability of the dynamics used for decision-making.

\subsection{Analysis of Learned Dynamics}
\label{sec:exp:qualitative}
The correctness of the learned and verified dynamics is classified as either correct or erroneous, with errors stemming from confounders, in-domain hallucinations, or out-of-domain hallucinations. As shown in Table \ref{tab:dynamics_cases}, an example of a confounder-related mistake occurs in Crafter when a simultaneous increase in health points is incorrectly attributed to the act of defeating a zombie. In this scenario, the Discoverer misclassifies the health increase as a direct result of defeating the zombie. In the case of in-domain hallucinations, the Discoverer incorrectly associates an increase in wood with defeating the zombie, even though it is not possible for wood to increase during this event, despite its presence in the observation. Lastly, out-of-domain hallucinations involve the discovery of dynamics that reference nonexistent objects in the observation or elements not present in the Crafter environment.

\begin{table}[t!]
\centering
\begin{tabular}{>{\raggedright\arraybackslash}p{4.8cm} >{\raggedright\arraybackslash}p{1.5cm}}
\toprule
\textbf{Correctness} & \textbf{Outcome} \\
\midrule
Correct ($\checkmark$) &  none \\
Confounder (\xmark) &  1 health \\
In-domain Hallucination (\xmark) &  1 wood \\
Out-domain Hallucination (\xmark) &  1 bone \\
\bottomrule
\end{tabular}
\caption{The dynamics underlying the outcome of defeating zombie in the Crafter environment.}
\label{tab:dynamics_cases}
\vspace{-0.5em}
\end{table}
Compared to the dynamics from the game manual, as shown in Table \ref{tab:dynamics_comparision}, we found that \methodname's dynamics are not only more precise but also more detailed. For instance, while SPRING only identified that placing a stone requires stones, \methodname determined that it specifically requires exactly one stone and the precise facing condition needed for successful placement. Moreover, using this information, the Evolver can infer advanced dynamics for placing a stone, such as its potential to serve as a barrier between the agent and dangerous creatures.

\begin{table}[h!]
\centering
\begin{tabular}{>{\raggedright\arraybackslash}p{1.5cm} >{\raggedright\arraybackslash}p{5cm}}
\toprule
\textbf{Sources} & \hspace{1.5cm} \textbf{Dynamics} \\
\midrule
Manual & Place stone requires stones \\
Discoverer & Place stone requires \textbf{1} stone and \textbf{faces paths, grass, sand, water, and lava} \\
Evolver  & Place stone to block zombies and skeletons, preventing them from reaching the player \\
\bottomrule
\end{tabular}
\caption{Comparing SPRING and \methodname on place stone's dynamics in the Crafter environment.}
\label{tab:dynamics_comparision}
\vspace{-1em}
\end{table}
In a customized MiniHack Lava Crossing setting, we modified the task such that it requires the agent to use the Wand of Death to freeze the lava for crossing, even though the wand’s original purpose is to zap monsters. Based on the demonstration, the Discoverer mistakenly identifies the precondition for using the wand as \minihack{The player must be adjacent to a wall or obstacle that can be destroyed or altered by the wand of death}, with the outcome being \minihack{The wand of death was used, altering the environment by removing walls and revealing a dark area to the east, southeast, and south}. 
This unreliable dynamic is successfully identified and filtered out by the Verifier, highlighting its crucial role in maintaining the reliability of learned dynamics.
\label{sec:exp}

\section{Related Work}
\paragraph{Language Models} Language models (LLMs) are trained autoregressively in a left-to-right sequence, predicting each token based on its preceding context from an internet-scale corpus. Through this training, LLMs develop a comprehensive understanding of both language and the world it represents~\cite{achiam2023gpt, touvron2023llama, dubey2024llama, TheC3}, enabling them to perform competently across a wide range of tasks~\cite{yao2024tree, shinn2024reflexion}.

\paragraph{Embodied Agent} Building an embodied agent using LLMs is challenging because LLMs lack embodied experience~\cite{valmeekam2022large, wang2024can} in downstream environments~\cite{weir2022one, shridhar2020alfworld, cote2019textworld}. However, LLMs can still provide foundational world knowledge that serves as a prior for the agent~\cite{shi2024opex, colas23augmenting, zhong2024policy, fu2024language}. A notable distinction of our work is that we do not assume LLMs have the necessary knowledge to solve tasks in specific domains.

\paragraph{Discover Dynamics} LLMs can discover knowledge by inducing and deducing rules for reasoning tasks~\cite{zhu2023large} and by extracting underlying domain knowledge from prior trajectories and interactive experiences~\cite{colas23augmenting, majumder2023clin, fu2024autoguide, zhao2024expel}. However, the knowledge obtained through these methods is often unstructured, not suited for addressing long-horizon planning problems, and lacks verification for reliability, as it overlooks LLMs' tendency to hallucinate~\cite{zhang2023hallucination}.

\paragraph{Evolve Dynamics} LLMs can refine their decision-making process by reflecting on past trajectories~\cite{shinn2024reflexion}. Leveraging this ability, studies such as \cite{wang2023voyager, stengel2024regal, zhang2023bootstrap, nottingham2024sso} focus on evolving new and advanced skills from pre-defined ones. However, these approaches often assume that LLMs already possess comprehensive domain knowledge and can derive new dynamics based solely on their understanding.

\section{Conclusion}
We introduce \methodname, a framework that bridges the knowledge gap between LLMs and downstream domains. The Discoverer extracts world dynamics, while the Verifier filters unreliable candidates. In an online setting, the Evolver reasons strategies through interaction. Our experiments demonstrate that \methodname effectively bridges the knowledge gap between LLMs and downstream domains.

\section*{Limitations}
Evaluating \methodname in embodied environments may not fully capture the diversity and complexity of real-world dynamics that an agent might encounter in practice. If the environment's dynamics change across episodes, the offline-learned dynamics from \methodname may struggle to adapt, introducing potential biases in the agent’s understanding and hindering the development of advanced strategies. Moreover, acquiring these dynamics from human demonstrations can be difficult or impractical in certain scenarios.

\section*{Ethical Concerns}
We do not anticipate any immediate ethical or societal impact from our work. This study aims to bridge the knowledge gap between LLMs and the target domain. However, despite our efforts, \methodname may still exhibit hallucinations due to the inherent tendency of LLMs to hallucinate.
\section*{Acknowledgements}
This work is supported by the Canada CIFAR AI Chair Program and the Canada NSERC Discovery Grant (RGPIN-2021-03115).

\clearpage
\bibliography{reference}

\clearpage
\appendix

\onecolumn

\hypersetup{
    colorlinks=true,
    linkcolor=black,         % Set link color to black
    urlcolor=black,
    citecolor=black
}

\renewcommand{\contentsname}{Appendix}
\setcounter{tocdepth}{0}
\tableofcontents

\setcounter{tocdepth}{2}
\addtocontents{toc}{\protect\setcounter{tocdepth}{2}}
\clearpage

\section{MiniHack}

This section offers further details on the MiniHack environment, along with a comprehensive description of the prompts utilized by \methodname within this environment. 

\subsection{Configuration}
\label{Appendix:minihack_setup}
For this environment, we employ the gpt-4o-2024-08-06 model.

\subsection{Des-Files for Each MiniHack Task}
In this subsection, we present the Des-Files utilized for the Lava Crossing, Wand of Death, and Quest tasks.
\label{appendix:des_files}
\subsubsection{Lava Crossing}
\begin{lstlisting}
MAZE: "mylevel", ' '
FLAGS:hardfloor
INIT_MAP: solidfill,' '
GEOMETRY:center,center
MAP
-------------
|.....L.....|
|.....L.....|
|.....L.....|
|.....L.....|
|.....L.....|
-------------
ENDMAP
REGION:(0,0,12,6),lit,"ordinary"
$left_bank = selection:fillrect (1,1,5,5)
$right_bank = selection:fillrect (7,1,11,5)
OBJECT:('=',"levitation"),rndcoord($left_bank),blessed
STAIR:rndcoord($right_bank),down
BRANCH:(1,1,5,5),(0,0,0,0)
\end{lstlisting}

\subsubsection{Wand of Death}
\begin{lstlisting}
MAZE: "mylevel", ' '
FLAGS:hardfloor
INIT_MAP: solidfill,' '
GEOMETRY:center,center
MAP
|---------------------------|
|...........................|
|.....|---------------------|
|.....|
|.....|
|-----|
ENDMAP
REGION:(1,1,5,5),lit,"ordinary"  
REGION:(6,1,26,1),lit,"ordinary"  
REGION:(26,1,27,1),lit,"ordinary"  
$safe_room = selection:fillrect(1,1,5,5)
OBJECT:('/',"death"),(1,1),blessed
MONSTER:('H', "minotaur"), (26,1)
STAIR:(27,1),down
BRANCH:(1,1,5,5),(1,1,2,2)
\end{lstlisting}

\subsubsection{Quest}
\begin{lstlisting}
MAZE: "mylevel", ' '
FLAGS:hardfloor
INIT_MAP: solidfill,' '
GEOMETRY:center,center
MAP
-------------| 
|.....L......|--      |-----|
|.....L........|------|.....|
|.....L.....................|
|.....L........|------|.....|
|.....L......|--      |-----|
-------------|
ENDMAP
REGION:(0,0,28,6),lit,"ordinary"
$left_bank = selection:fillrect (1,1,5,5)
$right_bank = selection:fillrect (7,1,11,5)
$goal_room = selection:fillrect (25,2,27,4)

OBJECT:('/',"cold"),(2,3),blessed  
BRANCH:(1,1,3,3),(0,0,0,0)
MONSTER:random, (25,3)
STAIR:(26,3),down
\end{lstlisting}

\subsection{Prompts for MiniHack Tasks}
This section includes the prompts used to solve the MiniHack tasks.

\subsubsection{Offline Dynamics Learning}
The prompts for offline learning the dynamics from demonstrations.

\noindent \textbf{Discover Action Dynamics}
\begin{lstlisting}
The player is playing a MiniHack game and would like you to analyze the game's dynamics. Specifically, they need help identifying the general preconditions and outcomes of actions by comparing consecutive observations.

- In the previous state, the player observed: {previous_state['state_description']} and took the action: {action}.
- In the current state, the player observed: {state['state_description']}.

Definitions:
- **General precondition**: The condition that must be met for the action to succeed, regardless of the specific game state.
- **General outcome**: The effect of the action, regardless of the specific game state.

Here is the action-to-text mapping: {state['action_prompt']}.
\end{lstlisting}

\noindent \textbf{Discover Object Dynamics}
\begin{lstlisting}
The player is playing a MiniHack game and would like you to analyze the game's dynamics. Specifically, they need help identifying the general preconditions and outcomes of actions by comparing consecutive observations.

- In the previous state, the player observed: {previous_state['state_description']} and took the action: {action}.
- In the current state, the player observed: {state['state_description']}.

Definitions:
- **General precondition**: The condition that must be met for the action to succeed, regardless of the specific game state.
- **General outcome**: The effect of the action, regardless of the specific game state.

Here is the action-to-text mapping: {state['action_prompt']}.
\end{lstlisting}

\noindent \textbf{Discover Subtask Dynamics}
\begin{lstlisting}
The player is playing a MiniHack game and would like your help in discovering the detailed steps to complete a subtask, and the preconditions and outcomes of the subtask. The player has provided the following information:
- In previous states, the player observed: {previous_state_description} and took the following actions: {previous_action}.

Definition:
- **Steps for completing the subtask**: The general and detailed sequence of actions, represented in text, that the player must follow to achieve the subtask. These steps should be general and not tied to the specific state of the environment.
- **Preconditions**: The general conditions that must be met for the subtask to be completed.
- **Outcomes**: The general effects of completing the subtask.

Here is the mapping of actions to their text representations: {state['action_prompt']}.
\end{lstlisting}

\subsubsection{Online Dynamics Learning}

The prompts for online evolving the dynamics from existing dynamics.

\noindent \textbf{Discover Subtask Dynamics}
\begin{lstlisting}
You are playing the MiniHack game.
Your current subtask is: {subtask}.
Your current observation is: {observation}.
The primitive dynamics of the game are: {action_dynamics}.
The object dynamics of the game are: {object_dynamics}.

You are asked to evolve useful strategies that aid in completing the subtask based on the provided information.

Instructions for evolving strategies:
- Do not introduce any new objects that are not part of the primitive dynamics.
- The evolved dynamics should not contradict the primitive dynamics.
- If a difficulty cannot be resolved by existing dynamics, evolve new and advanced dynamics by combining only existing dynamics using deductive reasoning.

Now, consider:
- All the potential difficulties that may arise, based on the primitive dynamics and current observation.
- All the potential dangers that may be encountered, based on the primitive dynamics and current observation.
- The strategies should be general that do not include specific actions, directions and dis


List all the potential difficulties, dangers and the evolved strategies to resolve them.
\end{lstlisting}

\subsubsection{Grounding}
The prompts for grounding the agent.

\noindent \textbf{Task Termination}
\begin{lstlisting}
You are playing the MiniHack game.
Your observation is: {description}.
Your current subtask is: {current_subtask}.
Your previous history is: {previous_history}.
The action and character mapping is: {action_prompt}.

You are asked to decide whether the current subtask should been terminated:
- If the termination conditions are met or the player is in danger or the player is in deadlock, the subtask should be terminated.

Output yes if the subtask has been completed, and no if it has not been completed.
\end{lstlisting}

\noindent \textbf{Task Termination}
\begin{lstlisting}
You are playing the MiniHack game.
Your observation is: {description}.
Your current subtask is: {current_subtask}.
Your previous history is: {previous_history}.
The action and character mapping is: {action_prompt}.

You are asked to decide whether the current subtask should been terminated:
- If the termination conditions are met or the player is in danger or the player is in deadlock, the subtask should be terminated.

Output yes if the subtask has been completed, and no if it has not been completed.
\end{lstlisting}

\noindent \textbf{Task Selection}
\begin{lstlisting}
You are playing the MiniHack game.
Your goal is: {task_description}.
Your observation is: {description}.
Your previously completed subtask is: {previous_subtask}.
All the subtasks are: {subtask_dynamics}.
All the environment dynamics are: {action_dynamics}.
All the object dynamics are: {object_dynamics}.

You are asked to select the next best subtask for completing the goal, based on all the provided information.
\end{lstlisting}

\noindent \textbf{Action Selection}
\begin{lstlisting}
You are playing the MiniHack game. 
Your current subtask is: {subtask}.
Your current observation is: {description}.
You also observe the feedback from the environment for the previous action: {observation}, which you need to consider for the next action.
The action and character mapping is: {action_prompt}.
Here are the strategies for the subtask: {evolved_strategy}.
Here are the primitive dynamics of the game: {action_dynamics} and {object_dynamics}.

Your past action and thought history: {history}
Now, you are asked to:
- List the potential difficulties and dangers that may arise based on the **current observation** and primitive dynamics for completing the subtask.
- Output the thoughts on the future actions how to resolve the difficulties and dangers for completing the subtask.
- Select the best action for the next step based on the thoughts.
- Do not copy the previous thoughts.

In the game, the distance is only measured as very far > far > near
\end{lstlisting}

\newpage

\section{Crafter}
This section provides a detailed overview of the Crafter environment and a comprehensive description of the prompts used by \methodname. 

\subsection{Configuration}
\label{Appendix:crafter_setup}
For this setup, we utilize the gpt-4o-2024-05-13 model in the Crafter environment with seeds 1, 2, 3, 4, and 5.

\subsection{An Example of Learned Dynamics}
\begin{lstlisting}
    (*@\textbf{grass}@*)
    grass can be found near ['tree', 'water', 'path'], but it is not associated with ['diamond', 'coal', 'iron']  
    You can walk directly through grass.  
    grass can only be used for: ['collect_sapling', 'eat_plant']  ,
    (*@\textbf{coal}@*)
    coal can be found near ['stone', 'iron', 'diamond'], but it is not associated with ['grass', 'cow', 'skeleton']  
    You cannot walk directly through coal.
    coal turn into path after collect_coal  
    coal can only be used for: ['make_iron_pickaxe', 'make_iron_sword', 'collect_coal']  
    coal can be collected by ['wood_pickaxe']  ,
\end{lstlisting}

\subsection{An Example of Trajectory}
\begin{lstlisting}
(*@\textbf{subgoal}@*)
Collect 4 units of wood and maintain a 
healthy level of health, food, drink, 
and energy.

(*@\textbf{subtask}@*)
collect_wood": {
    "General Plan": {
        "step_1": "Locate a tree, ensuring 'tree' is within immediate distance.",
        "step_2": "Move closer to the tree if it's not within immediate distance. This may involve navigating around obstacles such as grass, stone, or paths.",
        "step_3": "Face the tree to meet the precondition of the action.",
        "step_4": "Execute the 'collect_wood' action to gather wood from the tree."
    },
    "termination_condition": "The inventory's wood amount increases by 1, or if health, food, drink, or energy levels drop critically low, or a threat (such as skeletons or zombies) is detected nearby."
}

(*@\textbf{state description}@*)
It is daytimeState description: 
- East: immediate (grass); nearby (unexplored_area); 
- North: immediate (grass); nearby (unexplored_area); 
- North East: immediate (grass); nearby (unexplored_area); 
- North West: immediate (grass); nearby (unexplored_area); 
- South: immediate (grass); nearby (unexplored_area); 
- South East: immediate (grass); nearby (cow, unexplored_area); 
- South West: immediate (grass); nearby (unexplored_area); 
- West: immediate (grass); nearby (unexplored_area); 
Closest:
- grass: North 1 blocks away (immediate) (objects in between: None) 
- cow: South East 5 blocks away (nearby) (objects in between: {'grass'}) 
- Facing grass on the south.
Your status:
- health: 9/9
- food: 9/9
- drink: 9/9
- energy: 9/9
You have nothing in your inventory.

(*@\textbf{evolve dynamics}@*)
Expand the exploration area to discover new surroundings that could potentially include a tree.

(*@\textbf{action}@*)
move_north
\end{lstlisting}

\clearpage

\subsection{Changes in Dynamics Precision Over Time}
\begin{figure}[h!]
\centering
\includegraphics[width=0.5\textwidth]{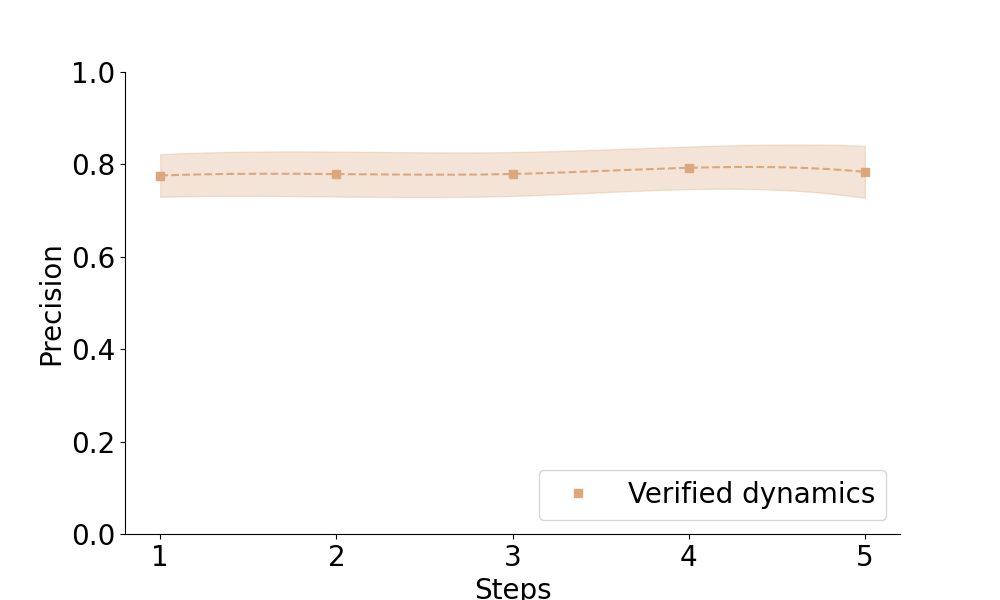}
\caption{Precision of verified dynamics over verified steps}
\label{fig:dynamic_precision_over_steps}
\vspace{-1em}
\end{figure}
As shown in Figure~\ref{fig:dynamic_precision_over_steps}, the precision of the verified dynamics $\mathcal{W}$ increases significantly during the initial verification step. However, as the process continues, further improvements in precision are minimal, indicating that the first step successfully filters out most of the unreliable dynamics.

\subsection{An Example of Verbalized Observation}
\label{sec:verbalizer_example}
\begin{figure}[h!]
\centering
\includegraphics[width=0.4\textwidth]{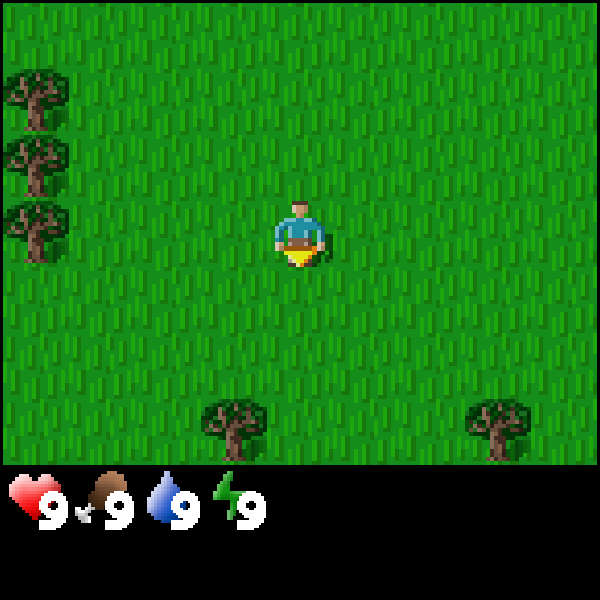}
\caption{The observation within the Crafter environment}
\label{fig:observation}
\end{figure}
\vfill
\newpage
\begin{lstlisting}
    (*@\textbf{Verbalized Observation}@*)
    It is daytime
    State description: 
    - East: immediate (grass); nearby (unexplored_area); 
    - North: immediate (grass); nearby (unexplored_area); 
    - North East: immediate (grass); nearby (unexplored_area); 
    - North West: immediate (grass); nearby (tree, unexplored_area); 
    - South: immediate (grass); nearby (unexplored_area); 
    - South East: immediate (grass); nearby (unexplored_area, tree); 
    - South West: immediate (grass); nearby (tree, unexplored_area); 
    - West: immediate (grass); nearby (tree, unexplored_area); 
    Closest:
    - grass: North 1 blocks away (immediate) (objects in between: None) 
    - tree: West 4 blocks away (nearby) (objects in between: {'grass'}) 
    - Facing grass on the south.
    Your status:
    - health: 9/9
    - food: 9/9
    - drink: 9/9
    - energy: 9/9
    You have nothing in your inventory.

\end{lstlisting}
\clearpage

\subsection{Tech Tree in the Crafter Environment}
\label{tech_tree_section}
\begin{figure*}[ht!]
\centering
\includegraphics[width=0.7\textwidth]{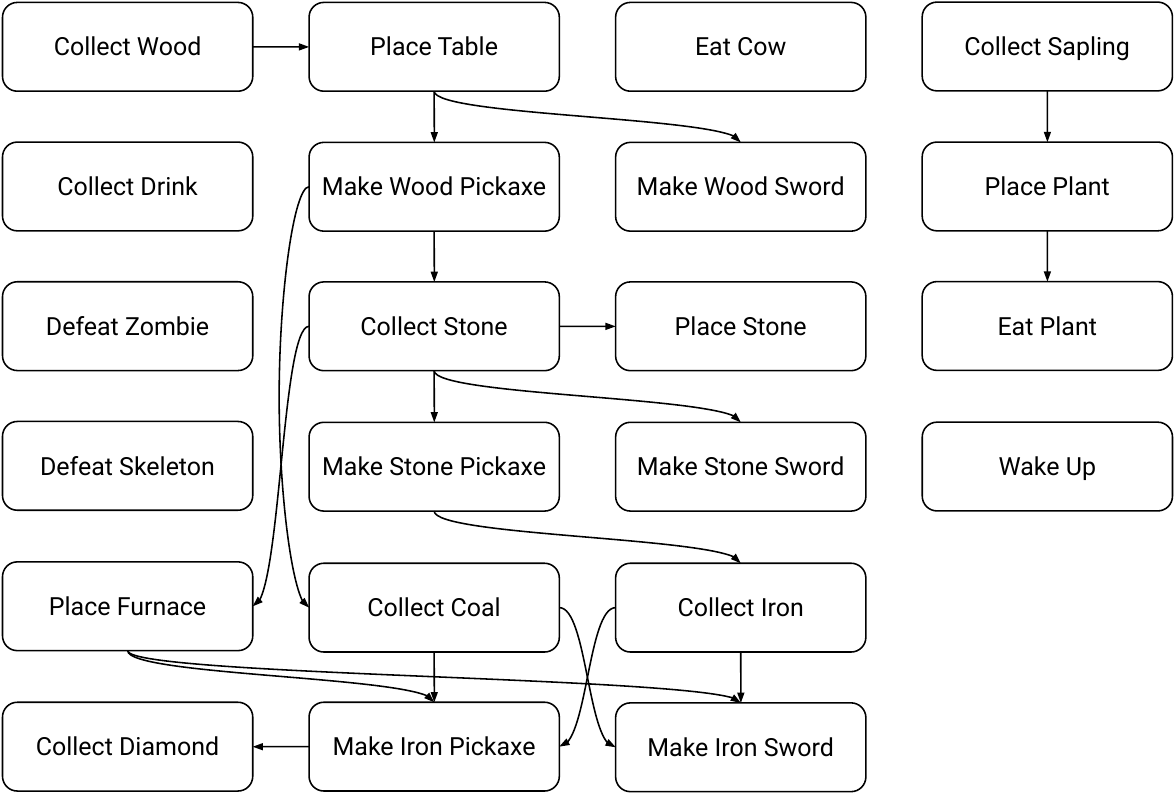}
\caption{The tech tree within the Crafter environment}
\label{fig:tech_tree}
\end{figure*}

\clearpage

\subsection{Prompts for Crafter Environment}
This section includes the prompts used in the Crafter environment.

\subsubsection{Online Dynamics Learning}
\begin{lstlisting}
Given the following details:
- Primitive dynamics: {env_dynamics}
- Current subtask: {transition['subtask']}
- Current observation: {transition['state_description']}

You are asked to identify the difficulties in completing the current subtask and provide 3 primitive or evolved dynamics to solve each of these difficulties.

Instructions for identifying difficulties:
- List the objects required to complete the subtask, specify their locations, and explain where to find them if they are not in the current observation. 
- Outline all possible obstacles that may be encountered along the way.

Instructions for evolving advanced dynamics:
- Do not introduce any new objects that are not part of the primitive dynamics.
- The evolved dynamics should not contradict the primitive dynamics.
- If a difficulty cannot be resolved by existing dynamics, evolve new and advanced dynamics by combining only existing dynamics using deductive reasoning.

Instructions for providing deductive reasoning steps:
- For each evolved dynamics, provide the used primitive dynamics.
- For the deductive reasoning steps, provide the steps to combine the primitive dynamics to evolve the advanced dynamics and the rule of inference used (Modus Ponens, Modus Tollens, ......)

Last, for each situation and dynamics should be general and do not contain details about specific locations about the objects.
Output in the following format:{output_format} and leaves 'None' for deductive_reasoning_steps if the dynamics are primitive.
\end{lstlisting}
\clearpage

\begin{lstlisting}
Given the following details:
- Primitive dynamics: {env_dynamics}
- Evolved dynamics: {transition['reformatted_dynamics']}
- Current subtask: {transition['subtask']}

You are asked to examine the validity of the evolved dynamics and provide feedback.

Instructions for examining the validity of the evolved dynamics:
- The evolved dynamics should only be a combination of existing dynamics using deductive reasoning and should not introduce new dynamics. Output 'True' if the evolved dynamics introduce new dynamics; otherwise, output 'False'.
- The evolved dynamics should not introduce any new objects that are not mentioned in the primitive dynamics. Output 'True' if the evolved dynamics introduce new objects; otherwise, output 'False'.
- Each deductive reasoning step should not contradict any of the primitive dynamics. Output 'True' if the evolved dynamics contradict the primitive dynamics; otherwise, output 'False'.

Instructions for examing the usefulness of the evolved dynamics:
- The difficulties should be directly related to the current subtask.
- The evolved dynamics should be useful in solving the difficulties identified in the subtask.
- Evluating the usefulness of the evolved dynamics on a scale of 1 to 5, where 5 is the most useful and 1 is the least useful.

Last, output the validity of the evolved dynamics in the following format: {output_format}.
\end{lstlisting}

\subsubsection{Grounding}
The prompts for grounding the agent.

\noindent \textbf{Subtask Termination}
\begin{lstlisting}
Given the following details:
- Subtask description: {self._transition['subtask']},
- Current observation: {self._transition['state_description']},
- Initial observation: {self._transition['initial_state_description']},
- Previous executed actions: {self._transition['previous_actions'][-3:]}
you are asked to decide whether the subtask should be terminated or not.

For deciding whether to terminate the subtask, consider:
- The previous action, provided it was executed successfully.
- The difference between the initial and current observations, including the inventory changes.

The subtask should be terminated, and the output should be 'True' only if any of its termination conditions are met. 
Otherwise, if none of the termination conditions are met, the subtask should continue running, and the output should be 'False'.

Justify whether the termination conditions are met or not first, and then provide the termination decision.
Output in this format: {output_format}.
\end{lstlisting}

\noindent \textbf{Action Selection}
\begin{lstlisting}
Given the following details:
- Current observation: {self._transition['state_description']}
- Current subtask's description: {self._transition['subtask']}
- Previous actions: {self._transition['previous_actions'][-3:]}
- Primitive dynamics: {self._transition['primitive_dynamics']}

You are asked to:
- identify the objects related to the current subtask and provide their locations and dynamics.
- select the top 3 actions that contributes to the subtask by either moving closer to the object or interacting with the object; and provide all the objects and dynamics directly related with each action.
- based on each action's related objects, provide the rationale and detailed consequences of executing each action on the objects.
- select the best action to execute next and provide the justification for your choice.

Lastly, select the action only from the available actions: {self._transition['available_actions']}; {feedback}.

Note: Avoid unnecessary crafting and placement if the items are within reachable distance.

Please format your response in the following format: {action_format}
\end{lstlisting}
\clearpage

\subsubsection{Offline Dynamics Learning}

\noindent \textbf{Learning Basic Attribute}
\begin{lstlisting}
In the Crafter environment, the world dynamics are unique. 

You are tasked with discovering the unique dynamics of actions within this environment.

Given the state transitions described as follows: {partial_description}, identify the materials used in the inventory for executing this action.

To discover the materials used in the inventory, consider the following aspects:
- List all the materials used not gained for each state transition and 
their corresponding quantities; 
- List all the common materials used in all state transitions and their corresponding 
quantities.

If no materials are used in common, simply output as "None".
materials are: wood, stone, coal, iron, diamond, sapling.  
Output as many pre-conditions as you can think of in the following format: {output_format}
\end{lstlisting}

\noindent \textbf{Learning Basic Attribute}
\begin{lstlisting}
In the Crafter environment, the world dynamics are unique. 

You are tasked with discovering the unique dynamics of actions within this environment.

Given the state transitions described as follows: {partial_description}, identify the objects within immediate distance required for executing this action.

To discover the objects within immediate distance, consider the following aspects:
- List all the objects within immediate distance for each state transition before executing the action, which you can only choose from the object list.
- Identify and list the objects that are present in all state transitions within the immediate distance before executing the action, which you can only choose from the object list.
- The common object **must** be present in all state transitions' immediate objects.

If no objects within immediate distance are in common, simply output as "None".
Object list: [coal, cow, diamond, furnace, iron, lava, skeleton, stone, table, tree, water, zombie, plant]
Output as many pre-conditions as you can think of in the following format: {output_format}
\end{lstlisting}

\clearpage
\noindent \textbf{Learning Basic Attribute}
\begin{lstlisting}
In the Crafter environment, the world dynamics are unique. 

You are tasked with discovering the unique dynamics of actions within this environment.

Given the state transitions described as follows: {partial_description}, identify the facing object required for executing this action.

To discover the objects within immediate distance, consider the following aspects:
- List the facing object for each state transition before this action executed, not the direction.
- List the union of all the facing object across all the state transitions before this action executed.
- The union of all the facing object must be present in at least one state transitions' facing object.

Object list: [coal, cow, diamond, furnace, iron, lava, skeleton, stone, table, tree, water, zombie, plant]
Output as many pre-conditions as you can think of in the following format: {output_format}
\end{lstlisting}

\noindent \textbf{Learning Basic Attribute}
\begin{lstlisting}
In the Crafter environment, the world dynamics are unique. 

You are tasked with discovering the unique dynamics of actions within this environment.

Given the state transitions described as follows: {partial_description}, identify the the inventory tool required for executing this action.

To discover the objects within immediate distance, consider the following aspects:
- List all the tools in the inventory for each state transition.
- List the tools that are common across all the state transitions.
- The common tools must be present in all state transitions' tools; if no tools are required, simply output as "None".
- List the most advanced tool required for executing the action within the common tools.

Tool list: [None, wood pickaxe, stone pickaxe, iron pickaxe, wood sword, stone sword, iron sword].
Output as many pre-conditions as you can think of in the following format: {output_format}
\end{lstlisting}

\clearpage

\noindent \textbf{Learning Basic Attribute}
\begin{lstlisting}
In the Crafter environment, the world dynamics are unique. 

You are tasked with discovering the unique dynamics of actions within this environment.

Given the state transitions described as follows: {partial_description}, identify the inventory and status increase after executing this action.

To discover the increase about the inventory and status, consider the following aspects:
- List all the increase of the inventory and status for each state transition after executing the action.
- List the increase of the inventory and status that are common across all the state transitions after executing the action.
- The common increase must be present in all state transition's increase; if no inventory and status can be found, simply output as "None";

Output as many pre-conditions as you can think of in the following format: {output_format}
\end{lstlisting}

\noindent \textbf{Learning Basic Attribute}
\begin{lstlisting}
In the Crafter environment, the world dynamics are unique. 

You are tasked with discovering the unique dynamics of actions within this environment.

Given the state transitions described as follows: {partial_description}, identify the facing object changes for executing this action.

To discover the change about the facing object, consider the following aspects:
- List all the change about the facing object after executing the action for each state transition.
- List the common change about the facing object across all the state transitions.
- The common change must be present in all state transitions' facing object change; if no change can be found, simply output as "None".

inventory: [wood, stone, coal, iron, diamond, sapling, wood_pickaxe, stone_pickaxe, iron_pickaxe, wood_sword, stone_sword, iron_sword]
status: [health, food, drink, energy], the maximum value of status is 9 and you can use 'increased_to_9' to represent the status increase to the maximum value.
Output as many pre-conditions as you can think of in the following format: {output_format}
\end{lstlisting}

\clearpage
\noindent \textbf{Verifying Basic Attribute}
\begin{lstlisting}
    In the crafter environment, you have discovered unique dynamics and need to verify the pre-conditions for the action '{action}'.
    Given the state transitions before and after taking the action '{action}', described as follows: {sampled_descriptions} 
    
    You are asked to verify the inventory materials used pre-conditions for the action '{action}' based on the discovered dynamics.
    
    Precondition: You need to use {precondition} to execute the action.
    
    For the verification, you need to consider the followings:
    - List all the inventory materials that are consumed not gained for each state transition and their corresponding quantity, 
    - List the common inventory materials used across all the state transitions.
    - Determine if the inventory materials used mentioned in the pre-condition are within the common inventory materials used.
    - If it is within the common inventory materials used, then it is valid; otherwise, it is invalid.
    
    Finally, if there are more advanced tools exist, then it is invalid; otherwise, it is valid.

    materials are: wood, stone, coal, iron, diamond, sapling.
    Output in this format: {output_format}
\end{lstlisting}

\noindent \textbf{Verifying Basic Attribute}
\begin{lstlisting}
In the crafter environment, you have discovered unique dynamics and need to verify the pre-conditions for the action '{action}'.
Given the state transitions before and after taking the action '{action}', described as follows: {sampled_descriptions} 

You are asked to verify the immediate objects pre-conditions for the action '{action}' based on the discovered dynamics.

Precondition: You only need {precondition} within immediate distance.

For the verification, you need to consider the followings:
- List all the objects within immediate distance before executing the action for each state transition only from the object list.
- Identify and list the objects that are present in all state transitions within the immediate distance before executing the action as common immediate objects, only from the object list.
- The common object **must** be present in all state transitions' immediate objects.
- Determine if there are objects mentioned in the pre-condition that are not in the common immediate objects.
- If there are objects mentioned in the pre-condition that are not in the common immediate objects, then it is invalid; otherwise, it is valid.

Object list: [coal, cow, diamond, furnace, iron, lava, skeleton, stone, table, tree, water, zombie, plant]
Output in this format: {output_format}
\end{lstlisting}

\clearpage
\noindent \textbf{Verifying Basic Attribute}
\begin{lstlisting}
In the crafter environment, you have discovered unique dynamics and need to verify the pre-conditions for the action '{action}'.
Given the state transitions before and after taking the action '{action}', described as follows: {sampled_descriptions} 

You are asked to verify the facing object pre-conditions for the action '{action}' based on the discovered dynamics.

Precondition: You need to face {precondition} before execute the action.

For the verification, you need to consider the followings:
- List the facing object for each state transition before this action executed.
- List the union of all the facing object across all the state transitions before this action executed.
- Determine if the facing object mentioned in the pre-condition is within the union of all the facing object.
- If the facing object mentioned in the pre-condition is within the union of all the facing objects, then it is valid; otherwise, it is invalid.

Finally, if there are more advanced tools exist, then it is invalid; otherwise, it is valid.
Object list: [coal, cow, diamond, furnace, iron, lava, skeleton, stone, table, tree, water, zombie, plant]
Output in this format: {output_format}
\end{lstlisting}

\noindent \textbf{Verifying Basic Attribute}
\begin{lstlisting}
In the crafter environment, you have discovered unique dynamics and need to verify the pre-conditions for the action '{action}'.
Given the state transitions before and after taking the action '{action}', described as follows: {sampled_descriptions} 

You are asked to verify the inventory pre-conditions for the action '{action}' based on the discovered dynamics.

Precondition: You only need {preconditions} to execute the action.

For the verification, you need to consider the followings:
 List all the tools in the inventory for each state transition.
- List the tools that are common across all the state transitions.
- The common tools must be present in all state transitions' tools; if no tools are required, simply output as "None".
- If the tools mentioned in the pre-condition are within the common tools, then it is valid; otherwise, it is invalid.

Tool list: [None, wood pickaxe, stone pickaxe, iron pickaxe, wood sword, stone sword, iron sword].
Output in this format: {output_format}
\end{lstlisting}
\clearpage

\noindent \textbf{Verifying Basic Attribute}
\begin{lstlisting}
In the crafter environment, you have discovered unique dynamics and need to verify the outcome for the action '{action}'.
Given the state transitions before and after taking the action '{action}', described as follows: {sampled_descriptions} 

You are asked to verify the inventory and status increase after executing this action. '{action}' based on the discovered dynamics.

predicted_increases: You only increase {precondition} in the inventory and status after executing the action.

For the verification, you need to consider the followings:
- List all the increases of the inventory and status for each state transition after executing the action as the increases.
- List the increases of the inventory and status that are common across all the state transitions after executing the action as the common increases.
- The common increases must be present in all state transition's increases; if no inventory and status increases can be found, simply output as "None".
- Determine if there are increases mentioned in the predicted_increases that are not in the common increases.
- If there are increases mentioned in the predicted_increases that are not in the common increases, then it is invalid; otherwise, it is valid.
Output in this format: {output_format}

inventory: [wood, stone, coal, iron, diamond, sapling, wood_pickaxe, stone_pickaxe, iron_pickaxe, wood_sword, stone_sword, iron_sword]
status: [health, food, drink, energy], the maximum value of status is 9 and you can use 'increased_to_9' to represent the status increase to the maximum value from the non-maximum value.
\end{lstlisting}

\noindent \textbf{Verifying Basic Attribute}
\begin{lstlisting}
In the crafter environment, you have discovered unique dynamics and need to verify the outcome for the action '{action}'.
Given the state transitions before and after taking the action '{action}', described as follows: {sampled_descriptions} 

You are asked to verify the facing object changes after executing this action. '{action}' based on the discovered dynamics.

predicted_changes: The facing object changes is {precondition} after executing the action.

For the verification, you need to consider the followings:
- List all the change about the facing object after executing the action for each state transition.
- List the common change about the facing object across all the state transitions.
- The common change must be present in all state transitions' facing object change; if no change can be found, simply output as "None".
- Determine if the predicted changes are within the common changes.
- If the predicted changes are within the common changes, then it is valid; otherwise, it is invalid.
\end{lstlisting}

\clearpage

\noindent \textbf{Verifying Basic Attribute}
\begin{lstlisting}
 Given randomly selected frames featuring '{obj}', please complete the following:
1. Identify which objects are most closely related to '{obj}'.
- Rate their relevance using these terms: 'Very related', 'Not related'.
- Here are the selected frames concerning '{obj}': {description}
2. Identify the common time relationship for this object.
3. Format your response as follows: {object_relationship_format}

Only consider the relationship with the following objects: ['grass', 'coal', 'cow', 'diamond', 'iron', 'lava', 'skeleton', 'stone', 'tree', 'water', 'zombie', 'plant', 'path', 'sand', 'plant-ripe']
For the relationship definitions:
- Very related: They can always be found together.
- Not related: They cannot be found together.
\end{lstlisting}

\noindent \textbf{Verifying Basic Attribute}
\begin{lstlisting}
Given randomly selected frames featuring '{obj}' described as: {description}, please complete the following tasks:

1. Identify the objects and time most closely related to '{obj}' in the provided description.
    - Rate their relevance using the following terms: 'Very related, 'Not related'.
    
2. Based on the discovered dynamics {dynamics}, verify the relationships between the objects and time.

3. Output all the valid relationships within the discovered dynamics that match all the relevance levels of the objects in the provided frames only about the object '{obj}'.

Consider relationships with the following objects only: ['grass', 'coal', 'cow', 'diamond', 'furnace', 'iron', 'lava', 'skeleton', 'stone', 'table', 'tree', 'water', 'zombie', 'plant', 'path', 'sand', 'plant-ripe'].

Relationship definitions:
- Very related: They are always found together.
- Not related: They are never found together.

Output in the following format: {object_relationship_format}
\end{lstlisting}

\clearpage

\noindent \textbf{Discover Subtask Attribute}
\begin{lstlisting}
In the Crafter environment, the world dynamics are complex and there are many subtasks to be completed.
Here are the dynamics about the core action {action}:
The precondition:
- it need to face: {self.facing_object_preconditions[action]}
- it need to have {self.immediate_object_preconditions[action]} within immediate distance
- it need to have {self.inventory_materials_precondition[action]} in the inventory.
- it need to have {self.inventory_tool_precondition[action]} in the inventory.
The outcome:
- its facing object changes to {self.facing_object_change[action]}
- its inventory and status outcome increases on {self.inventory_outcome[action]}
Note: 'None' indicates that there are no specific preconditions or outcomes for the corresponding elements of the dynamics."

Given the subtask "{subtask}", I want you to write the step-plan for completing this subtask.
First, read and understand all the provided world dynamics.
Then, locate the world dynamics that are relevant to the subtask.
Next, write the subtask's requirements, the step-plan for completing the subtask based on the world dynamics and the outcomes in this format {subtask_plan_format}
-steps: the general steps required to complete the subtask.
-termination_condition: when should this subtask be terminated; you should consider the status, potential danger and the outcome of the action
Finally, only output the plan for the subtask.
Here are a few examples: {examples}

Here are the observations on how to completing the subtask: {partial_description}
\end{lstlisting}

\noindent \textbf{Verify Subtask Attribute}
\begin{lstlisting}
In the Crafter environment, the world dynamics are complex and there are many subtasks to be completed.
Here are the dynamics about the core action {action}:
The precondition:
- it need to face: {self.facing_object_preconditions[action]}
- it need to have {self.immediate_object_preconditions[action]} within immediate distance
- it need to have {self.inventory_materials_precondition[action]} in the inventory.
- it need to have {self.inventory_tool_precondition[action]} in the inventory.
The outcome:
- its facing object changes to {self.facing_object_change[action]}
- its inventory and status outcome increases on {self.inventory_outcome[action]}
Note: 'None' indicates that there are no specific preconditions or outcomes for the corresponding elements of the dynamics."

Given the subtask "{subtask}", I want you to verify the plan for completing this subtask {discovered_plan}.
First, read and understand all the provided world dynamics.
Then, locate the world dynamics that are relevant to the subtask.
Next, examine the provided plan's requirements, the step-plan and the termination conditions.
Lastly, if it can be applied across different examples, then it is a valid plan.
Here are the observations on how to completing the subtask: {partial_description}
Output within this format{subtask_verification_format}
\end{lstlisting}

\clearpage

\noindent \textbf{Discover Subgoal Attribute}
\begin{lstlisting}
Given the following considerations:
- The game's final goal: collect diamonds and survive.
- All the subtasks: {['collect_coal', 'collect_diamond', 'collect_drink', 'collect_iron', 'collect_sapling', 'collect_stone', 'collect_wood', 'defeat_skeleton', 'defeat_zombie', 'eat_cow', 'eat_plant', 'make_iron_pickaxe', 'make_iron_sword', 'make_stone_pickaxe', 'make_stone_sword', 'make_wood_pickaxe', 'make_wood_sword', 'move', 'place_furnace', 'place_plant', 'place_stone', 'place_table', 'sleep']}
- Environment dynamics: {env_dynamics}
- Human player completed subtasks in chronological order: {human_demo}
Note: action_x_y means this action is performed consecutively for y times

Your task is to generate a plan by reordering the subtasks to help the player achieve the final game goal with different subgoals.

For plan generation, consider:
- List all the subtasks that contributes to the final goal from the human player's trajectory.
- Re-order the subtasks that fits the game's dynamics.
- For each subtask, state it as a subgoal with a description and the exact repetition of the subtask if needed.

Lastly, output the plan in this format: {output_format}; 

For example: {output_example}
\end{lstlisting}

\noindent \textbf{Verify Subgoal Attribute}
\begin{lstlisting}
Given the following considerations:
- The game's final goal: collect diamonds and survive.
- All the subtasks: {['collect_coal', 'collect_diamond', 'collect_drink', 'collect_iron', 'collect_sapling', 'collect_stone', 'collect_wood', 'defeat_skeleton', 'defeat_zombie', 'eat_cow', 'eat_plant', 'make_iron_pickaxe', 'make_iron_sword', 'make_stone_pickaxe', 'make_stone_sword', 'make_wood_pickaxe', 'make_wood_sword', 'move', 'place_furnace', 'place_plant', 'place_stone', 'place_table', 'sleep']}
- Environment dynamics: {env_dynamics}
- Human player completed subtasks in chronological order: {human_demo}
Note: action_x_y means this action is performed consecutively for y times
- Discovered plan: {discovered_plan}

Your task is to verify the provided plan. 

For plan verification, consider:
- List all the subtasks that contributes to the final goal from the human player's trajectory.
- Re-order the subtasks that fits the game's dynamics.
- If the provided plan can be applicable to the different trajectories, then it is valid.

Lastly, output this format: {output_format}; 

\end{lstlisting}

\end{document}